\documentclass[runningheads]{llncs}
\usepackage[T1]{fontenc}
\usepackage{graphicx}
\usepackage{hyperref}       
\usepackage{url}            

\usepackage{color}

\usepackage[utf8]{inputenc} 
\usepackage[T1]{fontenc}    
\usepackage{booktabs}       
\usepackage{amsfonts}       
\usepackage{nicefrac}       
\usepackage{microtype}      
\usepackage{xcolor}         

\usepackage{amsmath}
\usepackage{amssymb}
\usepackage{multirow}
\usepackage{makecell}
\usepackage{arydshln}
\usepackage[capitalize]{cleveref}

\usepackage{caption}
\usepackage{subcaption}
\usepackage[export]{adjustbox}

\def\AA{AutoAttack}
\begin{document}
\title{Diffusion or Non-Diffusion Adversarial Defenses: Rethinking the Relation between\\Classifier and Adversarial Purifier}
%
\author{Yuan-Chih Chen \and
Chun-Shien Lu}
\institute{Institute of Information Science, Academia Sinica, Taipei, Taiwan, ROC\\
\email{\{willpower057, lcs\}@iis.sinica.edu.tw}\\
}
%
\maketitle              

\begin{abstract}
Adversarial defense research continues to face challenges in combating against advanced adversarial attacks, yet with diffusion models increasingly favoring their defensive capabilities.
Unlike most prior studies that focus on diffusion models for test-time defense, we explore the generalization loss in classifiers caused by diffusion models.
We compare diffusion-based and non-diffusion-based adversarial purifiers, demonstrating that non-diffusion models can also achieve well performance under a practical setting of non-adaptive attack.
While non-diffusion models show promising adversarial robustness, they particularly excel in defense transferability and color generalization without relying on additional data beyond the training set.
Notably, a non-diffusion model trained on CIFAR-10 achieves state-of-the-art performance when tested directly on ImageNet, surpassing existing diffusion-based models trained specifically on ImageNet.
\end{abstract}    
\section{Introduction}
\label{sec:intro}
Proliferation of deep learning models across various domains has raised a pressing concern: Vulnerability of these models to adversarial attacks \cite{athalye2018BPDA,croce2020AutoAttack,kurakin2018BIM,madry2018PGD} that aim to make the model behave abnormally by manipulating the input data with imperceptible perturbations.
In response to these threats, researchers have actively investigated techniques to enhance the robustness of machine learning models against adversarial attacks in two branches: 
(1) Adversarial training ({\em e.g.},  \cite{hsiung2023towards,huang2023boosting,shafahi2019AdversarialTraining,wangICML2023,you2023NoisyImageModeling,wu2020adversarial}) 
involves both the clean and adversarially perturbed data for model training to improve robustness. 
While numerous studies have explored adversarial training, a notable disparity (from \href{https://robustbench.github.io/}{RobustBench}) persists between the natural accuracy and robust accuracy. 
(2) Adversarial purification ({\em e.g.},  \cite{alfarra2022AntiAdversaries,ho2022disco,nie2022DiffPure,song2024mimicdiffusion,zhang2023ScoreOpt}) takes a different approach by removing adversarial perturbations from input data prior to classification.
The benefit of adversarial purification is that it eliminates the need to retrain the classifier and can generalize to different attacks at test time.

Although diffusion models exhibit strong generalization across different attacks—known as ``attack generalization''—diffusion-based adversarial purifiers rely on pre-trained diffusion models to map images back to the training data distribution. However, this comes at the cost of reduced ``classifier generalization,'' making them sensitive to image processing like color variation and limiting their transferability across datasets. 
This limitations arise due to the differences in data augmentation and training objectives between the classifier and diffusion model.

Unlike most previous studies focusing on enhancing the effectiveness of cleansing adversarial images with diffusion models, we examine the classifier generalization loss linked to diffusion purifiers in this paper. 
To explore this issue, 
we consider the practical scenario of non-adaptive attack setting and different input variations. 
Our findings reveal that a non-diffusion purifier, trained using purification loss, can mitigate this generalization loss. Furthermore, we explain the underlying reasons for this phenomenon and highlight the differences between diffusion-based purifiers and non-diffusion-based purifiers.

The main contributions of this paper include:
\begin{itemize}
    \item
    Unlike most prior studies that focus on enhancing the performance of diffusion-based adversarial purifiers, our work is the first to investigate the classifier generalization loss induced by such purifiers.
    \item
    We explain why diffusion-based purifiers degrade classifier accuracy when processing images with slight variations from the training data, whereas purification-loss-based purifiers, a kind of non-diffusion-based purifiers, better preserve classification performance.
    \item We observe that diffusion-based purifiers are particularly sensitive to color variations. To investigate this phenomenon, we propose ColoredImageNet, a modified ImageNet to evaluate the impact of color shifts on purification effectiveness.
    
\end{itemize}

\section{Related works}

In this section, we review recent advancements in adversarial purification methods, followed by an exploration of how Masked Autoencoders (MAE) \cite{he2022MAE} have been applied to address the problems of demanding robustness.

Anti-Adv \cite{alfarra2022AntiAdversaries} introduces an anti-adversary layer designed to steer the image $x$ away from the decision boundary. 
The perturbation direction is guided by the image prediction via classifier, given the absence of true labels during inference. 
However, this strategy depends on the adversarial image being classified correctly—an assumption that may not always hold—potentially leading to misdirected corrections.
DISCO \cite{ho2022disco} uses the concept of LIIF \cite{chen2019LIIF} to extract the per-pixel feature by a pre-trained EDSR \cite{lim2017EDSR}.
Its training focuses solely on purification loss. DISCO demonstrates both acceptable robust accuracy and strong model transferability across different datasets.
DiffPure \cite{nie2022DiffPure} employs a diffusion model for image purification and provides a theoretical guarantee: 
By introducing sufficient Gaussian noises in the forward process, adversarial perturbations can be effectively eliminated. 
Regardless of the classifiers or attacks, DiffPure remains effective with the caveat that the diffusion timestep must strike a balance. Actually, it should be large enough to remove adversarial perturbations yet small enough to preserve global label semantics.
Building on DiffPure, ScoreOpt \cite{zhang2023ScoreOpt} introduces score-based priors into the purification process. Adversarial samples are optimized to converge toward regions with higher posterior likelihood, as defined by pre-trained score-based models.
More recently, MimicDiffusion \cite{song2024mimicdiffusion} diverges from DiffPure’s approach. Instead of adding noise to adversarial images, it starts from pure Gaussian noise and applies a reverse diffusion process guided by the adversarial input to generate purified images.



On the other hand, MAE \cite{he2022MAE} implements a masking mechanism to enhance the performance of ViT \cite{dosovitskiy2020ViT}. Inspired by the Masked Language Modeling (MLM) technique used in BERT \cite{kenton2019bert}, MAE operates as a pre-training model, focusing on learning patch representations during the pre-training stage and fine-tuning for downstream tasks. 
Hereafter, MAE and MLM will be interchangeably used.

Recently, some works \cite{huang2023FrequencyMAE,tsai2023testMAE,you2023NoisyImageModeling,wuDMAE} have employed MAE for the problems of demanding robustness.
Huang \textit{et al.} \cite{huang2023FrequencyMAE} and DMAE \cite{wuDMAE} focused on robustness in the context of classification tasks, rather than on image purification.
DRAM \cite{tsai2023testMAE} proposes a test-time detection method to repair adversarial samples in that the MAE reconstruction loss is directly used to detect the adversarial samples due to the assumption of different distributions of clean and adversarial samples.
NIM-MAE \cite{you2023NoisyImageModeling} uses the MAE structure to achieve adversarial training by injecting the noise into the entire image instead of masking patches within an image.



\section{Difference between Diffusion-Based Purifiers and Purification-Loss-Based Purifiers}
\label{sec:defense_generalization}

In this section, we first introduce the preliminary and notation in \cref{sec:Notation}, followed by an analysis of the accuracy drop caused by diffusion-based purifiers in \cref{sec:discrepancy_diffusion}. 
We examine the effectiveness of purification loss in Sec. \ref{sec:discussion_purification_loss} that will be an important component in our proposed non-diffusion-based purifier.

\subsection{Preliminary and Notation}
\label{sec:Notation}
Frequently used notations are as follows: Clean image $x$ and its corresponding label $y$; adversarial image $x_a$; classifier $c$, which outputs the predicted label $\Tilde{y}$; an adversarial purifier $\mathcal{P}$; an image $x \in \mathbb{R}^{H \times W}$ cropped into $N$ patches of area $ps\times ps$ (w.r.t. patch size $ps$) before forwarding to purifier; purifier $\mathcal{P}_{MAE} = g \circ f$ with MAE encoder $f$ and MAE decoder $g$; masking ratio $r$ of MAE and corresponding binary mask $M$; and ${\bf 1}$ is a matrix with all elements of $1$.

\subsubsection{Adversarial Attack.}
Given a classifier $c$ parameterized by $\theta$, and a clean data pair $(x, y)$, an adversarial attack aims to find an adversarial sample $x_a$ derived from $x$ to deceive the classifier ({\it i.e.,} $c(x_a) \neq y$) by the optimization as: $\max_{x_a} \ L(x_a, y;\theta)$, s.t. $||x_a-x||_p<\epsilon$,
where $L$ is the training loss function, $||\cdot||_p$ denotes $p$-norm, and $\epsilon$ means the attack budget. 

\subsubsection{Generalized Purifier.} 
\label{sec:Preliminary:Purifier} 
Given a classifier $c$ and a clean data pair $(x, y)$, a purifier $\mathcal{P}$ aims to refine $x_a$ to prevent misclassification ({\it i.e.,} ensuring $c(\mathcal{P}(x_a)) = c(x)$) while preserving clean accuracy ({\it i.e.,} $c(\mathcal{P}(x)) = c(x)$). In this work, we further emphasize that the purifier should maintain the classifier's generalization capability, meaning it should also correctly process unseen images, particularly common real-world images ({\it i.e.,} $c(\mathcal{P}(x')) = c(x')$), where $x'$ represents such unseen data.

\subsubsection{Diffusion Model.}
DDPM \cite{ho2020DDPM} consists of a $T$-step forward process and a corresponding $T$-step reverse process. 
The forward process gradually corrupts a clean image $x_0 \sim q(x)$ by adding Gaussian noise, resulting in a noisy image $x_T$. The reverse process then aims to reconstruct the original image by denoising $x_T$ back to $x_0$.
The forward process is defined as:
\begin{align} 
q(x_t \mid x_0) = \mathcal{N}(x_t; \sqrt{\bar{\alpha}}x_0, (1-\bar{\alpha}_t)\mathbf{I}),
\label{Eq:diffusion_forward}
\end{align}
where $t$ is the diffusion timestep, $\alpha_t$ is a predefined noise schedule, and $\bar{\alpha}_t = \prod_{i=0}^{t} \alpha_i$ represents the cumulative product of the noise schedule. The distribution $q(x)$ represents the clean training data distribution.
The reverse process approximates the posterior $q(x_{t-1} \mid x_t)$ by iteratively sampling:
\begin{align} 
p_{\theta}(x_{t-1} \mid x_t) = \mathcal{N}(x_{t-1}; \mu_{\theta}(x_t,t), \sigma_t \mathbf{I}),
\end{align}
where $\mu_{\theta} =  \frac{1}{\sqrt{\alpha_t}}(x_t - \frac{\beta_t}{\sqrt{1-\bar{\alpha}_t}}\epsilon_\theta (x_t, t))$, $\sigma_t \mathbf{I}$ is untrained time dependent constant, and $\epsilon_{\theta}$ denotes the approximate noise predicted by the neural network. 
To train the model $\epsilon_\theta$, a clean image $x_0 \sim q(x)$ is randomly selected, and Gaussian noise $\epsilon \sim \mathcal{N}(0, \mathbf{I})$ is added at a random timestep $t$. The model is trained to predict the noise component using the following objective:  
\begin{align} 
\nabla_{\theta} || \epsilon - \epsilon_{\theta}( \sqrt{\bar{\alpha}}x_0 + \epsilon \sqrt{1-\bar{\alpha}} ) ||^2_2.
\end{align} 

\subsubsection{Masked Autoencoder.}
Given an image ${x} \in \mathbb{R}^{H \times W}$ and a binary mask $M \in \mathbb{R}^{H \times W}$, the goal of MAE \cite{he2022MAE} is to reconstruct the entire image $\Bar{x} = x\in \mathbb{R}^{H \times W}$ from partial image ${M \odot x}$ with the reconstruction loss defined as:
\begin{align} 
L_{MAE}(x, \Bar{x})=\|({\bf 1}-M) \odot \Bar{x} - ({\bf 1}-M) \odot g \circ f(M \odot x)\|_2,
\label{Eq:MAE}
\end{align}
where ${\bf 1}$ is a matrix equal to $1^{H \times W}$, $\odot$ is the element-wise product of two matrices of the same size, $f$ is MAE encoder, $g$ is MAE decoder, and $M$ is a random binary image mask parameterized by the masking ratio $r$ as $||M||_1 = (1 - r)\times(H\times W)$.
The masking ratio $r=0$ if there is no masking.

\subsection{The Discrepancy between Classifier and Diffusion Purifier}
\label{sec:discrepancy_diffusion}

The predominant purification-based approaches \cite{nie2022DiffPure,song2024mimicdiffusion,zhang2023ScoreOpt} in adversarial defense involve the diffusion model, which primarily modifies the reverse diffusion process of DiffPure \cite{nie2022DiffPure} using the same pre-trained model. 
However, these approaches primarily use clean accuracy to demonstrate that the purifier preserves the classifier's ability to classify non-attacked images. 
Different from prior studies, we find that directly applying a pre-trained diffusion model can actually be detrimental to the classifier—an issue that cannot be detected solely through clean accuracy.

\begin{figure*}[t]
    \centering        
        \begin{subfigure}{0.3\textwidth}
            \centering
            \includegraphics[width=.9\linewidth]{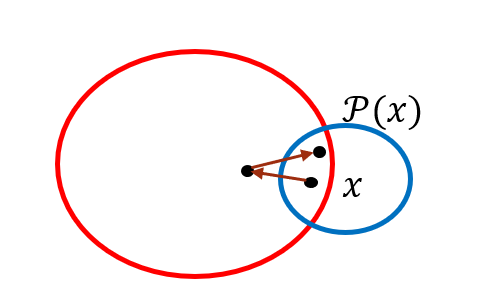}
            \caption{}\label{fig:diffusion_illustration1}            
        \end{subfigure}        
        \begin{subfigure}{0.3\textwidth}
            \centering
            \includegraphics[width=.9\linewidth]{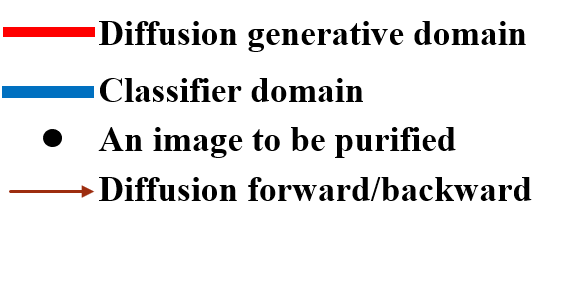}    
            \caption*{}\label{fig:111}      
        \end{subfigure}
        \begin{subfigure}{0.3\textwidth}
            \centering
            \includegraphics[width=.9\linewidth]{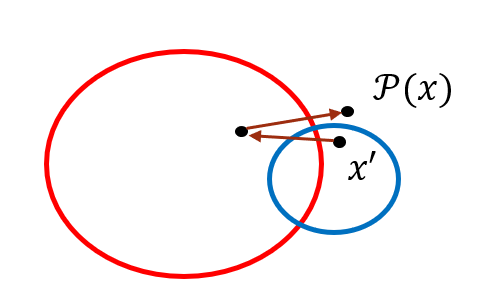}
            \caption{}\label{fig:diffusion_illustration2}
        \end{subfigure}

    \caption{Illustration of the discrepancy between the classifier and diffusion-based purifier. (a) When the input image $x$ is from the training set or a similar distribution (e.g., test data split from the same dataset), it has a higher likelihood of maintaining clean accuracy, as shown in previous studies. (b) However, if the input image $x'$ is an unseen sample that the classifier can correctly identify, diffusion-based purifiers are more prone to misclassification.}
    \label{fig:diffusion_illustration}
\end{figure*}

As illustrated in \cref{fig:diffusion_illustration}, the generative domain of a diffusion model (in red circle) differs from the classification domain of a classifier (in blue circle). The intersection of both circles contains the shared training data for both models, while the outer region of blue circle indicates the areas where mis-classification occurs. 
\cref{fig:diffusion_illustration}(a) depicts scenarios commonly used in previous studies to evaluate clean accuracy, whereas \cref{fig:diffusion_illustration}(b) illustrates how a purifier can degrade classifier performance when the test data slightly deviates from training data. 

To verify the scenario in \cref{fig:diffusion_illustration}(a), we present experiments in Sec. \ref{sec:exp_robustness}. 
As for \cref{fig:diffusion_illustration}(b), we analyze the effect of color variations in Sec. \ref{sec:Exp:Color_Transform} and common color corruptions in Sec. \ref{sec:Exp:Corruption} to support our claim.
Furthermore, we investigate a scenario in that the training data for purifiers and classifiers do not overlap ({\em i.e.}, the red and blue circles have no intersection). 
For this scenario, we explore in Sec. \ref{sec:Exp:transfer_dataset}, where a purifier is applied to a dataset different from its training dataset, and in Sec. \ref{Sec: More Transfering}, where we examine the transition from a low-resolution dataset to a high-resolution one.


To analyze the aforementioned scenarios, we define a test image that slightly deviates from the training data distribution as follows: 
\begin{align} 
x' = y = Ax + n,
\label{Eq:corrupted_image}
\end{align}
where $x$ is a clean image sampled from the training distribution $q(x)$, $A$ is a transformation operator, $n$ is additive Gaussian noise, and $y$ is the resulting transformed (or corrupted) image. 
When $A$ is the identity matrix and $n=0$, the transformation reduces to the original image, {\em i.e.}, $y = x$.
The forward process of the diffusion-based purifier $\mathcal{P}$ can then be formulated by combining \cref{Eq:diffusion_forward,Eq:corrupted_image} as 
$q(x_t \mid y) = \mathcal{N}(x_t; \sqrt{\bar{\alpha}_t}y, (1-\bar{\alpha}_t)\mathbf{I})$,
where the purifier operates using a reduced number of diffusion steps, typically with $t \in \{0.1, 0.15\}$, in contrast to the full diffusion process ($t = T$) employed in standard DDPMs.
Notably, when $t = T$, the diffusion purifier becomes equivalent to a full DDPM, since $q(x_T \mid y) \sim \mathcal{N}(0, \mathbf{I})$, leading to $\mathcal{P}(y) \sim q(x)$. Conversely, when $t = 0$, the purifier simply returns the input, {\em i.e.}, $\mathcal{P}(y) = Ax + n$.

\begin{figure*}[t]
    \centering        
    \includegraphics[width=.6\linewidth]{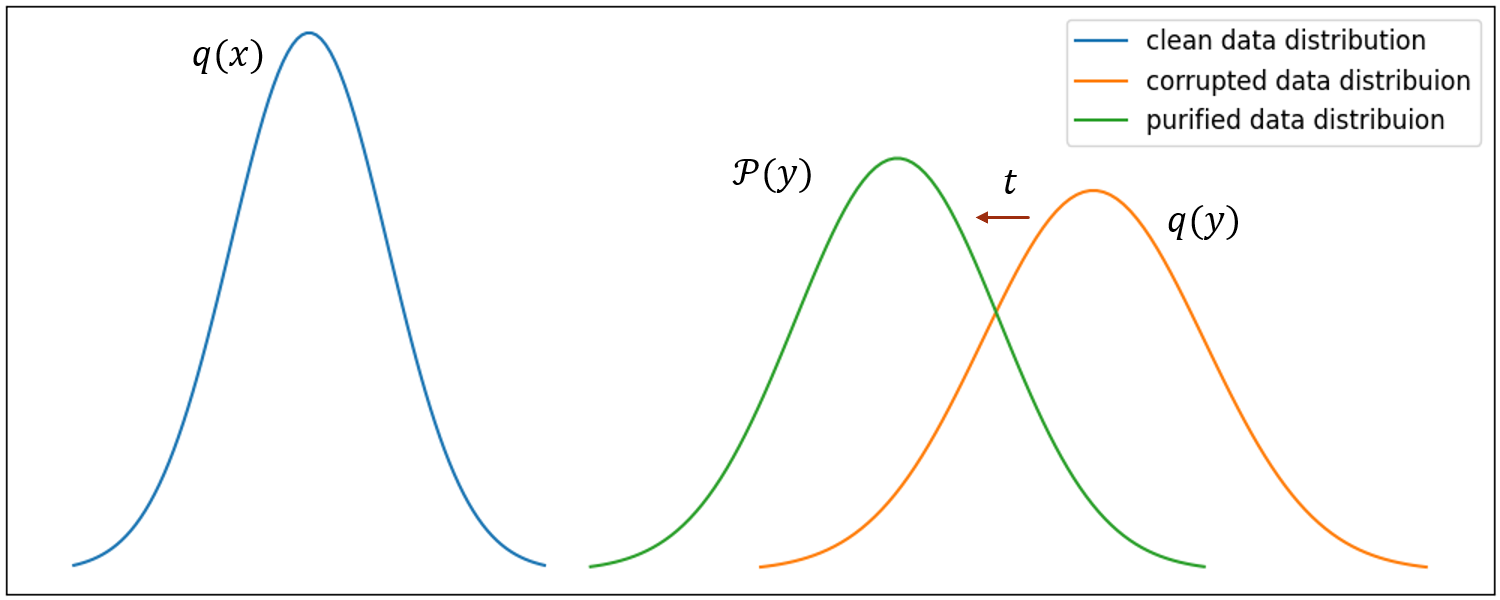} 
    \caption{Visualization of the purified image distribution across various timesteps $t$.}
    \label{fig:diffusion_shift}
\end{figure*}

According to Theorem 3.1 in \cite{nie2022DiffPure}, the relationship between the distributions $q(y)$ and $q(x)$ during the diffusion process satisfies:
\begin{align} 
\frac{\partial D_{KL}{(q(y)||q(x))}}{\partial t} \leq 0,
\end{align}
indicating that the KL divergence between them decreases monotonically as $t$ increases from 0 to 1. Consequently, the purified distribution $\mathcal{P}(y)$ transits from $q(y)$ toward $q(x)$, as illustrated in \cref{fig:diffusion_shift}.

In contrast to the ideal case of $A = I$ and $n=0$, 
applying a non-trivial transformation $A$ introduces a distributional shift. As a result, for $0 < t < 1$, the purified image $\mathcal{P}(y)$ deviates from $q(y)$, 
potentially leading to a drop in clean accuracy $c(\mathcal{P}(y))$ due to the divergence between the purified output and the true corrupted input distribution.

Specifically, diffusion models tend to generate images similar to training data, as noted in \cite{somepalli2023digital_forgery}. In contrast, classifiers are trained with data augmentation to enhance generalization and prevent overfitting. Since diffusion models focus on generating natural images, common data augmentation techniques—such as color jitter, rotation, and mixup—are typically avoided to prevent the generation of unrealistic or unnatural images.
This fundamental difference leads to a discrepancy: classifiers leverage data augmentation to learn texture variations, improving their ability to classify unseen images, while diffusion models push unseen images toward the closest points in the training distribution, which may not align with the classifier's expectations.
As we will demonstrate later in Sec. \ref{sec:Exp:Color_Transform}, this mismatch in augmentation strategies results in accuracy drops when handling color variations.

\begin{figure*}[h]
    \centering        
        \begin{subfigure}{0.21\textwidth}
            \centering
            \includegraphics[width=\linewidth]{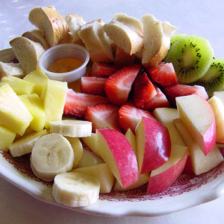}
            \caption{Clean image}\label{fig:purified_images_clean}
        \end{subfigure} \hspace{0.2in}
        \begin{subfigure}{0.21\textwidth}
            \centering
            \includegraphics[width=\linewidth]{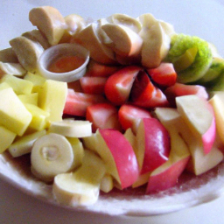}
            \caption{DiffPure}\label{fig:purified_images_diffpure}
        \end{subfigure} \hspace{0.2in}
        \begin{subfigure}{0.21\textwidth}
            \centering
            \includegraphics[width=\linewidth]{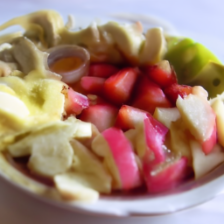}
            \caption{MimicDiffusion}\label{fig:purified_images_mimicdiffusion}
        \end{subfigure}

        \begin{subfigure}{0.21\textwidth}
            \centering
            \includegraphics[width=\linewidth]{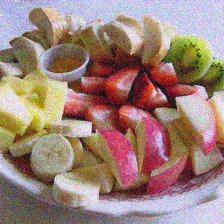}
            \caption{ScoreOpt-O}\label{fig:purified_images_score_x0}
        \end{subfigure} \hspace{0.15in}
        \begin{subfigure}{0.21\textwidth}
            \centering
            \includegraphics[width=\linewidth]{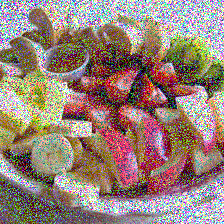}
            \caption{ScoreOpt-N}\label{fig:purified_images_score_xt_7}
        \end{subfigure} \hspace{0.15in}        
        \begin{subfigure}{0.21\textwidth}
            \centering
            \includegraphics[width=\linewidth]{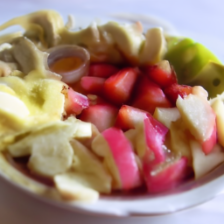}
            \caption{DDA}\label{fig:purified_images_dda_7}
        \end{subfigure} \hspace{0.15in}         
        \begin{subfigure}{0.21\textwidth}
            \centering
            \includegraphics[width=\linewidth]{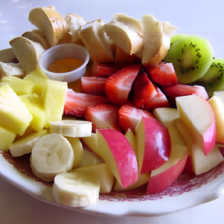}
            \caption{MAEP}\label{fig:purified_images_maep}
        \end{subfigure}  
        
    \caption{Illustration of purified clean images from ImageNet using diffusion-based purifiers (DiffPure \cite{nie2022DiffPure}, MimicDiffusion \cite{song2024mimicdiffusion}, ScoreOpt \cite{zhang2023ScoreOpt}, and DDA \cite{gao2023dda}) compared to the purification loss-based purifier (MAEP in Sec. \ref{sec:Method}):
    (a) The original, unmodified image from ImageNet.
    (b) DiffPure introduces significant alterations to image details and textures, with noticeable changes in the appearance of the kiwi, bread, and strawberry.
    (c) MimicDiffusion substantially modifies the structural composition of the image.
    (d-e) ScoreOpt, after the denoising process, still retains some noises.
    (f)  DDA, guided by a low-pass filter and designed to restore corrupted images to their natural states, still alters the image's texture.
    (g) MAEP applies minimal modifications, such as slight smoothing, to effectively purify the image while preserving its original structure.
    }
    \label{fig:purified_images}
\end{figure*}

To give a preliminary expression, as shown in \cref{Table:ssim_psnr_imagenet}, diffusion models exhibit lower SSIM and PSNR scores compared to non-diffusion-based purifiers (Proposed MAEP in Sec. \ref{sec:Method}), indicating a decline in image quality after purification.
Furthermonre, the purified images generated by diffusion models and those produced by the purifier trained with purification loss (MAEP) are shown in \cref{fig:purified_images}. 
It is evident that recent diffusion-based purifiers, such as DiffPure \cite{nie2022DiffPure}, ScoreOpt \cite{zhang2023ScoreOpt}, and MimicDiffusion \cite{song2024mimicdiffusion}, significantly alter image details during the purification process, whereas the purification loss-based approach effectively preserves more of the original image details while maintaining robustness. 
Simply altering the reverse process of diffusion-based purifiers by introducing randomness or estimation can significantly increase semantic loss in the image. For example, this effect is observed in ScoreOpt, as demonstrated later in Sec. \ref{sec:Exp:Corruption}.

While previous studies suggest that diffusion-based purifiers can maintain clean accuracy without necessarily preserving image quality, we argue that their limitations warrant further exploration. Specifically, these purifiers rely solely on the training dataset without access to classifier-specific information, which may impact their ability to effectively purify images. Additionally, the substantial alterations introduced during purification can lead to information loss for the classifier. This issue is often overlooked because hyperparameters are typically fine-tuned using validation accuracy metrics, inadvertently masking the impact of image quality degradation.


\begin{table*}[t]
  \centering
  \resizebox{0.4\textwidth}{!}{
  \begin{tabular}{@{}ccccc@{}}
    \toprule
    Defense Methods                              & PSNR ($\uparrow$) & SSIM ($\uparrow$) \\    
    \midrule[2pt]  
    DiffPure \cite{nie2022DiffPure}              & 25.50  & 0.73\\
    ScoreOpt-N \cite{zhang2023ScoreOpt}          & 15.99 & 0.31\\
    ScoreOpt-O \cite{zhang2023ScoreOpt}          & 22.01 &  0.52\\  
    MimicDiffusion \cite{song2024mimicdiffusion} & 22.06 & 0.68\\
    DDA \cite{gao2023dda}                        & 25.51 & 0.75\\  
    \hdashline
    MAEP (Sec. \ref{sec:Method})                                       &  \textbf{34.80} & \textbf{ 0.93} \\
    \midrule[2pt]   
  \end{tabular}}
  \caption{
  Evaluations of purifying images by non-diffusion based purifier (MAEP) and state-of-the-art diffusion-based methods.  
  Testing dataset: ImageNet.} 
  \label{Table:ssim_psnr_imagenet}
\end{table*}

\subsection{Purification Loss vs. Clean Accuracy}
\label{sec:discussion_purification_loss}
In the literature,   
DISCO \cite{ho2022disco} is found to achieve both acceptable clean and robust accuracy by employing just one purification loss, while preserving model transferability.
The purification loss is defined as:
\begin{align}    
L^{DISCO}_{purify}(x, \mathcal{P}(x_a)) = \ell_1(x, \mathcal{P}(x_a)),
\label{Eq:L1_purifier_loss}
\end{align}
where $\mathcal{P}$ is a purifier used to purify input image $x_a$ by reconstructing the clean image $x$ in terms of $\ell_1$-norm loss between $x$ and $x_a$.

Specifically, DISCO shows that it can purify the adversarial image efficiently, expressed as:
\begin{align}  
\mathcal{P}(x_a) \approx x,
\label{Eq:equivalent_adv}
\end{align}
\begin{align} 
c(\mathcal{P}(x_a)) \approx c(x),
\label{Eq:equivalent_adv2}
\end{align}
where Eq. (\ref{Eq:equivalent_adv}) denotes the perceptual similarity between the clean image $x$ and purified image $\mathcal{P}(x_a)$, Eq. (\ref{Eq:equivalent_adv2}) indicates label-preservation, and $c$ is a pre-trained classifier from \href{https://robustbench.github.io/}{RobustBench} or PyTorch official website.
Nevertheless, there is still a room for DISCO to improve label-preservation for purified clean images, defined as:
\begin{align}  
c(\mathcal{P}(x)) \approx c(x).
\label{Eq:equivalent_clean}
\end{align}
Although DISCO \cite{ho2022disco} does not ensure to preserve the clean accuracy in \cref{Eq:equivalent_clean}, 
it indeed shows good trade-off between the clean accuracy and robust accuracy in several testing scenarios, including different classifiers, different attack algorithms (such as Autoattack \cite{croce2020AutoAttack}, PGD \cite{madry2018PGD}, FAB \cite{croce2020fab}, BIM \cite{kurakin2018BIM}, BPDA \cite{athalye2018BPDA}, and FGSM \cite{goodfellow2015FGSM}), and transferability to different datasets.

Based on the above observations, we conjecture that the purification loss ($\ell_1(\mathcal{P}(x_a), x)$ in \cref{Eq:L1_purifier_loss}) can efficiently purify the adversarial image without remarkably sacrificing clean accuracy.
Here, we provide a feasible but simple explanation of why DISCO can have acceptable performance without needing to consider $\ell_1(\mathcal{P}(x), x)$, as illustrated in \cref{fig:plot}.

\begin{table}[t]
\centering
\begin{minipage}{.42\linewidth}
    \centering
        \includegraphics[width=0.8\linewidth]{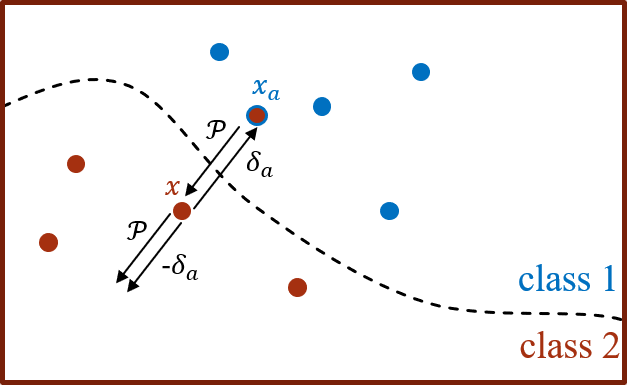}        
    \captionof{figure}{Purification loss learns the direction from $x_a$ to $x$ and the direction of $-\delta_a$ (one-step adversarial perturbation along negative gradient) is roughly the same as $\mathcal{P}(x_a)$ due to \cref{Eq:equivalent}.}
    \label{fig:plot}
\end{minipage}%
\hfill
\begin{minipage}{.54\linewidth}    
  \centering
  \resizebox{0.6\textwidth}{!}{
  \begin{tabular}{@{}ccc@{}}
    \toprule
           & Acc. (\%)  & Formula\\
    \midrule
    $c(\mathcal{P}(x_a))$ & 95.16  & \cref{Eq:equivalent_adv2} \\
    $c(\mathcal{P}(x))$   & 89.94  & \cref{Eq:vector_AntiAdv} \\
    \bottomrule
  \end{tabular}}
  \caption{\small Verification of our conjecture on the purification loss ($\ell_1(\mathcal{P}(x_a), x)$ in \cref{Eq:L1_purifier_loss}. For $c(\mathcal{P}(x)) \approx c(x-\delta_a)$, we pre-process the training dataset by adding $-\delta_a$ (via PGD) and feeding it to a non-defense classifier (ResNet-18) pre-trained on CIFAR-10 for testing. 
  Since our derivation introduces approximation to remove the influence of purifier and is only based on a classifier, it means that this is a theoretical accuracy of a purifier and \cref{Eq:L1_purifier_loss} can properly maintain clean accuracy.
  \label{Tab:ADV}}
\end{minipage}
\end{table}

Specifically, based on the assumption that an adversarial image is created to be similar to its clean counterpart in term of $L_\infty$-norm as:
\begin{align}  
x \approx x_a\ s.t.\ |x-x_a|_\infty<\epsilon,
\label{Eq:equivalent}
\end{align}
we can derive
\begin{align}  
\mathcal{P}(x)-x \approx \mathcal{P}(x_a)-x_a = -\delta_a,
\label{Eq:vector}
\end{align}
where the direction of purification, $\mathcal{P}(x)-x$, for clean image $x$ is similar to $\mathcal{P}(x_a)-x_a$ of an adversarial image due to the adversarial condition specified in \cref{Eq:equivalent}.

In practice, input images fall into two categories: adversarial images ($x_a$) and clean images ($x$).
The robust accuracy of purifier is tied to $x_a$ and the prediction of $x_a$ can be calculated by $c(\mathcal{P}(x_a)) \approx c(x)$, as detailed in \cref{Eq:equivalent_adv2}.
The clean accuracy of purifier, which DISCO didn't make a discussion, is related to $x$ and the prediction of $\mathcal{P}(x)$ can be derived from \cref{Eq:vector} as:
\begin{equation}
c(\mathcal{P}(x)) = c(x+(\mathcal{P}(x)-x)) \approx c(x+(\mathcal{P}(x_a)-x_a)) = c(x-\delta_a),
\label{Eq:vector_AntiAdv}
\end{equation}
where $\mathcal{P}(x_a)-x_a$ is roughly equal to $-\delta_a$, as illustrated in \cref{fig:plot}.
To verify the above conjecture, we present the results in \Cref{Tab:ADV}.

Overall, to better preserve image semantics while maintaining clean accuracy, we propose incorporating the purification loss (\cref{Eq:L1_purifier_loss}) into Masked Language Model (MLM) in Sec. \ref{sec:Method}.


\begin{figure*}[t]
    \centering        
        \includegraphics[width=0.9\linewidth]{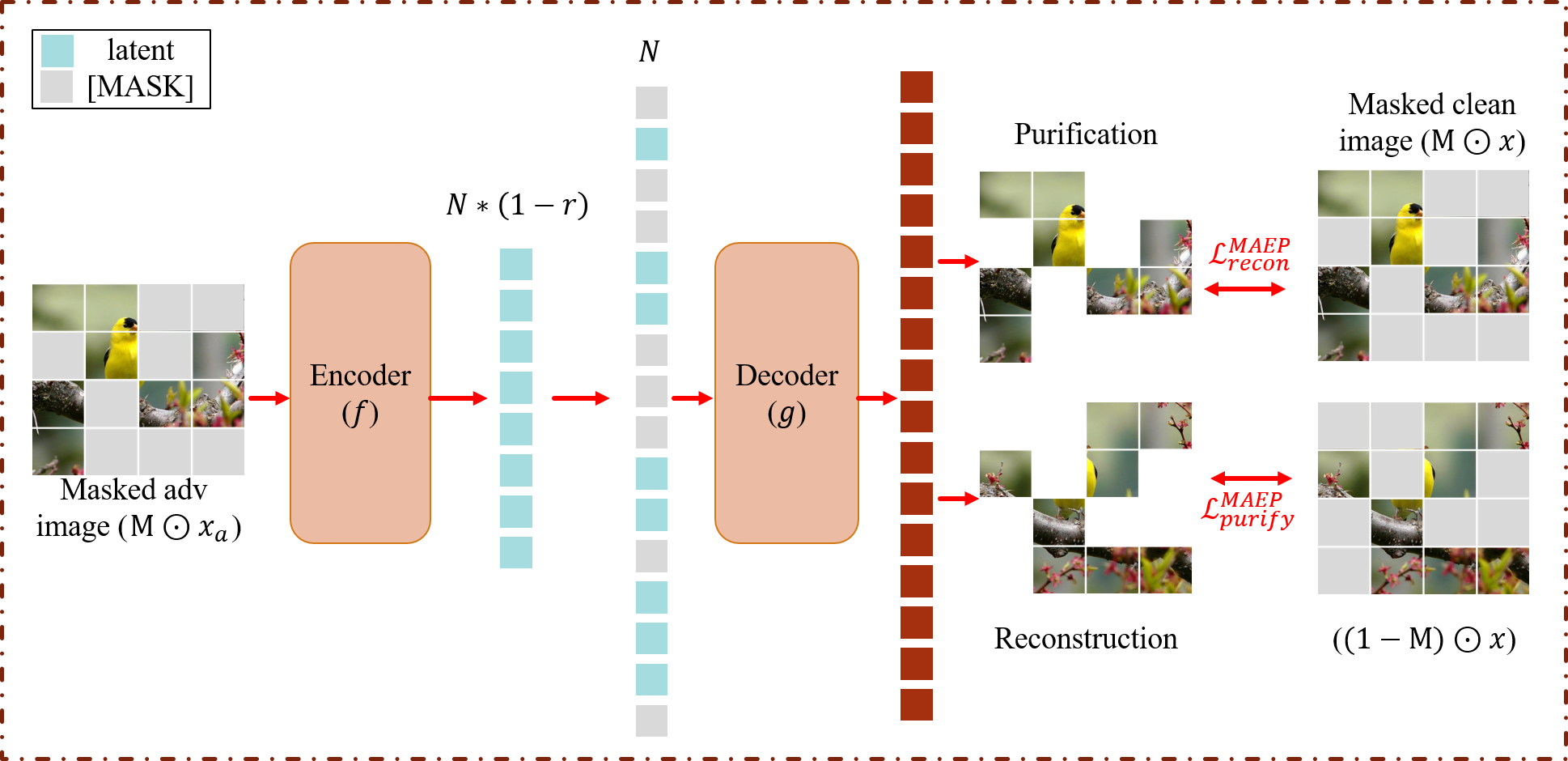}
    \caption{
     MAEP workflow: MAEP purifies adversarial images using purification loss. It then learns adversarial patch representations to further enhance purification while preserving semantic information by masking clean patch predictions.    
    }
    \label{fig:mil_flow}
\end{figure*}
\section{
Masked AutoEncoder Purifier (MAEP)}
\label{sec:Method}




Building on our findings in Sec. \ref{sec:discrepancy_diffusion} and Sec. \ref{sec:discussion_purification_loss}, MAEP leverages the masking mechanism to enhance purification by enforcing a purification loss that removes adversarial perturbations while preserving image semantics through a reconstruction loss.


\subsection{Objective Function Design}
\label{sec:ObjectiveFunction}

We study diverse objective function designs \cite{he2022MAE,ho2022disco,kenton2019bert,shafahi2019AdversarialTraining,zhang2019TRADES}, all geared towards enhancing robust accuracy. These designs include TRADES \cite{zhang2019TRADES}, MLM \cite{he2022MAE,kenton2019bert}, and reconstruction loss \cite{ho2022disco,shafahi2019AdversarialTraining}, where TRADES is a popular choice for training robust classifiers, MLM showcases effectiveness in vision tasks, and reconstruction loss is often used for image reconstruction. We will delve into a discussion of the results produced by these designs, emphasizing the superior performance of MAEP among these alternatives.
In the following, the distance measure $D$ is chosen to be $\ell_1$-norm as we find it performs better than mean square error.
The descriptions regarding reconstruction loss and extensions based on TRADES and MLM will be shown in Sec. \ref{Sec: Objective Functions} of Appendix.
In addition, the performance comparison of all loss designs can also be found in \Cref{Tab:objective_function}
of Sec. \ref{Sec: Objective Functions}.





\subsection{Total Loss in MAEP}
Unlike the aforementioned design, MAEP is investigated to integrate both the reconstruction loss of MLM and purification loss in \cref{Eq:L1_purifier_loss} to boost adversarial robustness and maintain the semantic of the image.
We also show that solely utilizing MAE/MLM does not yield satisfactory performance in Sec. \ref{Sec: Objective Functions} of Appendix.

The entire loss of MAEP can be separated into two parts. 
Part 1). The purification loss, adapted from \cref{Eq:L1_purifier_loss}, addresses the purification task focusing solely on the unmasked region within an image. 
Several reasons support this decision to exclusively handle the unmasked region.
First, training with partial image information can generalize to the entire image via position embedding, as outlined in MAE \cite{he2022MAE}. 
Second, the purifier is designed to operate on the entire image rather than the masked region. 
Third, the incorporation of MLM discussed in Part 2) below can effectively address the issue of dealing with the masked region.
Part 2). Unlike Part 1), which utilizes the unmasked region for purification, MLM reconstructs the masked region of an image using the unmasked portions. This masking mechanism helps the model learn adversarial representations and identify adversarial perturbations, thereby enhancing performance. Additionally, this loss ensures that MAEP preserves the semantic integrity of the image while maintaining the benefits of purification loss.

Based on the above concerns, it is ready to define the MAEP loss.
First, we define the purification loss.
Recall that $g \circ f$ is called the purifier in MAE.
To ensure the clean accuracy and robust accuracy, as described in Sec. \ref{sec:discussion_purification_loss}, we adopt  \cref{Eq:L1_purifier_loss} for masked region and the purification loss of unmasked region in MAEP is defined as:
\begin{equation}
L^{MAEP}_{purify}=\|M \odot x - M \odot g \circ f (M \odot x_a)\|.
\label{Eq:MAEP_purifier_loss}
\end{equation}
Note that $L^{MAEP}_{purify}$ reconstructs the clean image $x$ based on the adversarial image $x_a$, which is consistent with Part 1) and is different from the traditional MAE, as shown in \cref{Eq:MAE}.

Second, following the reconstruction loss ($L_{pre-train}$ in Eq. (\ref{Eq:Loss_MLM_pre-trin}) of Appendix) in MAE, the reconstruction loss $L^{MAEP}_{recon}$ of masked region in MAEP is defined as $L^{MAEP}_{recon} = L_{pre-train}$.
Thus, the entire loss of MAEP is derived as:
\begin{equation}
\begin{aligned}
L^{MAEP} &=L^{MAEP}_{purify} + L^{MAEP}_{recon}\\ 
&= \|M \odot x - M \odot g \circ f (M \odot x_a)\| + \|({\bf 1}-M) \odot x -  ({\bf 1}-M) \odot g \circ f (M \odot x_a)\|\\
&\geq \|x - g \circ f (M \odot x_a)\|,
\label{eq:total_loss}
\end{aligned}
\end{equation}
where the equal sign holds when $\|\cdot\|$ is $\ell_1$-norm, which is adopted as the distance measure in MAEP. 
The masking ratio $r \in (0,1)$ controls the image mask $M$.
\cref{eq:total_loss} will degenerate to the purification loss in  \cref{Eq:L1_purifier_loss} when $r=0$.

\section{Experiments}
\label{sec:Experiments}

In Sec. \ref{sec:defense_generalization}, we discuss the differences between diffusion-based and purification loss-based purifiers. 
To better preserve semantic information, we introduce MAEP, which enhances the training of purification-loss-based purifiers using a masked language modeling (MLM)-inspired objective, in Sec. \ref{sec:Method}. 
In this section, we first present robust accuracy results to demonstrate their competitive performance. We then conduct performance analysis under varying levels of discrepancy between training and inference data, from minor differences ({\em e.g.}, color variations) to more significant shifts ({\em e.g.}, different datasets), providing a clear understanding of the distinctions between non-diffusion-based purifiers and diffusion-based purifiers.


\subsection{Datasets, Model Settings, and Implementation Details}
\label{sec:Datasets_Detail}
Four commonly used datasets, CIFAR10 \cite{krizhevsky2009CIFAR10}, CIFAR100 \cite{krizhevsky2009CIFAR10}, ImageNet \cite{deng2009imagenet}, and ImageNet-C \cite{hendrycks2019imagenetC} were adopted.
For ColoredImageNet used here, it was generated by applying the method \cite{reinhard2001ColorTransfer} to transform ImageNet images to match the color styles of target images. The target images consist of $20$ samples from the ImageNet test set, resulting in a dataset that is $20$ times the size of the original ImageNet.
All models were trained using NVIDIA V100 GPUs.

For model architectures, we followed previous studies to employ WRN-28-10 \cite{zagoruyko2016WRN} and its corresponding model weights provided by \href{https://robustbench.github.io/}{RobustBench} for CIFAR10. 
However, for CIFAR100 and ImageNet, due to the absence of model weights from RobustBench, we adopted them from DISCO \cite{ho2022disco} and PyTorch, respectively.
For the attacks, we consider PGD-$\ell_\infty$ and AutoAttack, and set the permissible perturbation $\epsilon$ such that $| \epsilon |_\infty \le 8/255$.


To train the purifier, we first pre-trained MAEP from scratch using the loss function defined in \cref{eq:total_loss}, with a masking ratio of $r = 0.5$ and a patch size of 2. During inference, the masking ratio was set to $r = 0$ to fully utilize the input for downstream tasks.
The clean and robust accuracy results were averaged over five runs with different random seeds and followed the non-adaptive setting of DiffPure \cite{nie2022DiffPure}.

\begin{table*}[t]
  \centering
  \resizebox{1.0\textwidth}{!}{
  \begin{tabular}{@{}ccccc@{}}
    \toprule
    Defense Methods & Clean Accuracy (\%) & Robust Accuracy (\%) & Average Accuracy (\%) & Attacks\\
    \midrule[2pt]
    No defense & 94.78 & 0 & 47.39 & PGD-$\ell_\infty$/{\AA} (Standard)\\
    \hdashline
    AWP \cite{wu2020adversarial}*  & 88.25 & 60.05 & 74.15 & {\AA} (Standard)\\ 
    Anti-Adv \cite{alfarra2022AntiAdversaries}* + AWP  &88.25 & 79.21 & 83.73 & {\AA} (Standard)\\ \hdashline
    DISCO \cite{ho2022disco} & 89.26 & 82.99 & 86.12 & PGD-$\ell_\infty$\\
    DISCO \cite{ho2022disco}  & 89.26 & 85.33 & 87.29 & {\AA} (Standard)\\
    DiffPure \cite{nie2022DiffPure}  & 88.62 & 87.12 & 87.87 &PGD-$\ell_\infty$\\
    DiffPure \cite{nie2022DiffPure} & 88.15 & 87.29 & 87.72 &{\AA} (Standard)\\  
    ScoreOpt-N \cite{zhang2023ScoreOpt} & 91.03 & 80.04 & 85.53 &PGD-$\ell_\infty$\\
    ScoreOpt-O \cite{zhang2023ScoreOpt} & 89.16 & 89.15 & 89.15 &PGD-$\ell_\infty$\\	
    ScoreOpt-N \cite{zhang2023ScoreOpt} & 91.31 & 81.79 & 86.55 &{\AA} (Standard)\\
    ScoreOpt-O \cite{zhang2023ScoreOpt} & 89.18 & 89.01 & 89.09 &{\AA} (Standard)\\	
    SOAP* \cite{shi2021online} & 96.93 & 63.10 & 80.01 &PGD-$\ell_\infty$\\
    Hill {\em et al}. \cite{hill2020stochastic}* & 84.12 & 78.91 & 81.51 &PGD-$\ell_\infty$\\
    ADP ($\sigma=0.1$) \cite{yoon2021adversarial}* & 93.09 & 85.45 & \textbf{89.27} &PGD-$\ell_\infty$\\
    \hdashline
    MAEP  & 92.31 & 86.19 & 89.25 & PGD-$\ell_\infty$\\
    MAEP  & 92.30 & 88.73 & \textbf{90.52}& {\AA} (Standard)\\
    \midrule[2pt]   
  \end{tabular}}
  \caption{
  Robustness evaluation and comparison. 
  Classifier: WRN-28-10. 
  Testing dataset: CIFAR-10.
 Asterisk (*) indicates that the results were excerpted from the papers.}
   \label{Table:Robustness_cifar10}
\end{table*}
 
\begin{table*}[t]
  \centering
  \resizebox{1.0\textwidth}{!}{
  \begin{tabular}{@{}ccccc@{}}
    \toprule
    Defense Methods & Clean Accuracy (\%) & Robust Accuracy (\%) & Average Accuracy (\%) & Attacks\\
    \midrule[2pt]
    No defense & 81.66 & 0 & 40.83 & PGD-$\ell_\infty$/{\AA} (Standard)\\
    Rebuffi {\em et al.} \cite{rebuffiNeurIPS2021}  & 62.41 & 32.06 & 47.23 & {\AA} (Standard)\\
    Wang {\em et al.} \cite{wangICML2023}           & 72.58 & 38.83 & 55.70 & {\AA} (Standard)\\
    Cui {\em et al.} \cite{cuiarXiv2023}            & 73.85 & 39.18 & 56.51 & {\AA} (Standard)\\
    DISCO \cite{ho2022disco}      & 69.78 & 73.36           & \textbf{71.57}     &PGD-$\ell_\infty$\\
    DISCO \cite{ho2022disco}     & 69.78 & 76.91           & 73.34     & {\AA} (Standard)\\
    DiffPure \cite{nie2022DiffPure}  & 61.96 & 59.27 & 60.61 &PGD-$\ell_\infty$\\
    DiffPure \cite{nie2022DiffPure}  & 61.98 & 61.19 & 61.58 &{\AA} (Standard)\\
    \hdashline 
    MAEP            & 73.67 & 70.96 & \textbf{71.57} & PGD-$\ell_\infty$\\  
    MAEP            & 73.67 & 76.22 & \textbf{74.95} & {\AA} (Standard)\\
    \bottomrule
  \end{tabular}}
  \caption{Robustness evaluation and comparison.
  Classifier: WRN-28-10. Testing dataset: CIFAR-100.}
  \label{Table:Robustness_cifar100}
\end{table*}

\subsection{Evaluation of Adversarial Defense}
\label{sec:exp_robustness}
We adopted SOTA adversarial purifiers for comparison, including diffusion model-based approaches \cite{nie2022DiffPure,zhang2023ScoreOpt,yoon2021adversarial} and non-diffusion-based approaches \cite{alfarra2022AntiAdversaries,hill2020stochastic,ho2022disco,shi2021online,wu2020adversarial}.
For DiffPure \cite{nie2022DiffPure}, we used the official code and tested the purifier under the same experimental setup mentioned in Sec. \ref{sec:Datasets_Detail}. 
For ScoreOpt \cite{zhang2023ScoreOpt}, the classifier was trained by the authors and not from RobustBench. For a fair comparison, we used the official code and only replaced the default classifier of ScoreOpt with the WRN-28-10 model provided by RobustBench.

\Cref{Table:Robustness_cifar10} shows the comparison results with dataset CIFAR-10.
We have observations as follows:
(1) For robust accuracy, MAEP and ScoreOpt-O \cite{zhang2023ScoreOpt} are comparable but better than DiffPure \cite{nie2022DiffPure}. However, for clean accuracy, MAEP performs better than \cite{nie2022DiffPure}\cite{zhang2023ScoreOpt}.
(2) MAEP and diffusion-based defenses are generally better than other methods in terms of average accuracy.

For CIFAR100, the robustness comparison results are shown in \Cref{Table:Robustness_cifar100}.
Different from CIFAR-10, under CIFAR-100, DISCO performs better than DiffPure for both clean and robust accuracy, and MAEP outperforms DISCO and DiffPure with a large gap.
It is noteworthy that for both MAEP and DISCO, their robust accuracy is higher than clean accuracy. 
One possible explanation is that they primarily learn the mapping from the adversarial image $x_{adv}$ to the clean image $x$.
This conforms to the  verification in \Cref{Tab:ADV} and depicts that the theoretical optimal situation, in which the robust accuracy is higher than clean accuracy, may exist.

\begin{table}[t]
\centering
\begin{minipage}{.47\linewidth}
    \centering
    \includegraphics[width=.85\linewidth]{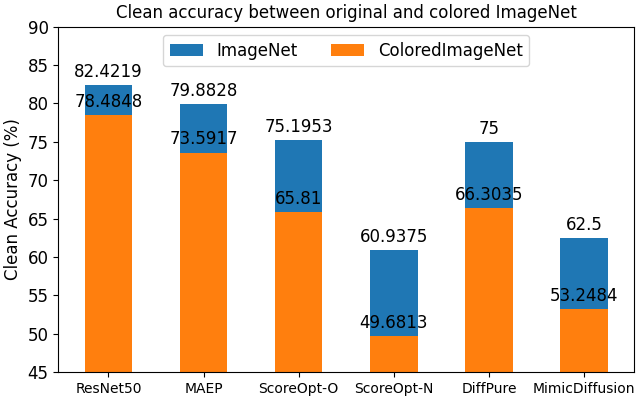}
    \caption{The illustration of clean accuracy drop across different purifiers on ColorImageNet.}
    \label{fig:color_imagenet_clean_acc}
\end{minipage}%
\hfill
\begin{minipage}{.49\linewidth}    
    \centering
    \includegraphics[width=.85\linewidth]{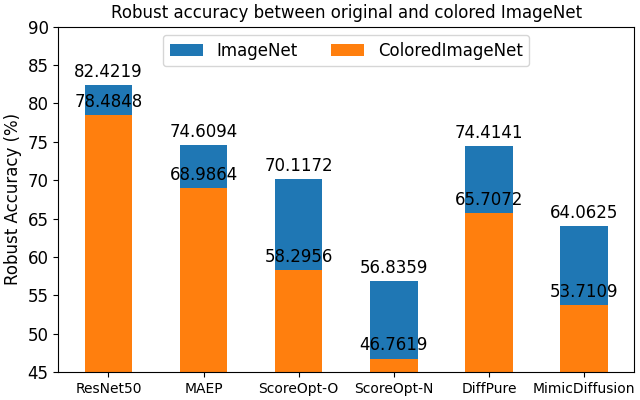}
    \caption{The illustration of robust accuracy drop across different purifiers on ColorImageNet.}
    \label{fig:color_imagenet_robust_acc}
\end{minipage}
\end{table}

\subsection{Sensitivity to Color Transform in Diffusion-based Purifiers}
\label{sec:Exp:Color_Transform}
As we argue in Sec. \ref{sec:Preliminary:Purifier}, the purifier should not degrade the classifier’s accuracy.
In this section, we aim to demonstrate that if a classifier learns the texture features of a class—such as a cat from a training dataset of brown cats, does it still correctly classify cats of different colors? 
To the best of our knowledge, however, no such dataset currently exists. Therefore, we generate ``ColoredImageNet'' using a color transfer technique  \cite{reinhard2001ColorTransfer}, as described in Sec. \ref{sec:Datasets_Detail}.
In Sec. \ref{sec:exp_robustness}, we have already demonstrated that non-diffusion and diffusion-based purifiers achieve comparable robustness. Here, we extend our verification to scenarios, where test images differ only in color, as illustrated in \cref{fig:color_imagenet_clean_acc,fig:color_imagenet_robust_acc}.
The results reveal that diffusion-based purifiers are more sensitive to color variations, as indicated by the difference between the blue bars for ImageNet and the orange bars for ColoredImageNet. Specifically, diffusion-based approaches—ScoreOpt \cite{zhang2023ScoreOpt}, DiffPure \cite{nie2022DiffPure}, and MimicDiffusion \cite{song2024mimicdiffusion}—experience an accuracy drop approximately twice as large as that of MAEP.


\begin{table}[t]
  \centering
  \resizebox{1.0\textwidth}{!}{
  \begin{tabular}{@{}lccccccc@{}}
    \toprule
    Model    & \multicolumn{2}{c}{Training Data} & \multicolumn{2}{c}{Test Data}  & Clean Acc. (\%) & Robust Acc. (\%) & Avg. Acc. (\%)\\
                & CIFAR10 & CIFAR100 & CIFAR10 & CIFAR100 & & \\    
    \midrule
    WRN28-10    & v & & v & & 94.78 & 0 & 47.39\\
    \midrule
     + DiffPure \cite{nie2022DiffPure} & v & & v & & 89.58  & 89.45 & 89.51\\
     + DISCO \cite{ho2022disco}   & v & & v & & 89.26  & 85.33 & 87.29\\     
     + MAEP     & v & & v & & 92.30  & 88.73 & \textbf{90.51}\\     
    \hdashline
    + DiffPure \cite{nie2022DiffPure} & & v & v & & 94.50  & 69.00 & 81.75\\
    + DISCO \cite{ho2022disco}   & & v & v & & 89.78  & 87.44 & \textbf{88.61}\\    
    + MAEP     & & v & v & & 91.58  & 84.73 & 88.16\\
    \bottomrule
  \end{tabular}}
 \caption{Transferability of adversarial defenses (DISCO, Diffpure, and MAEP) from CIFAR100 \cite{krizhevsky2009learning} to CIFAR10 \cite{krizhevsky2009learning} (trained on CIFAR100 but tested on CIFAR10) under WRN28-10 \cite{WRN} and  \AA \  \cite{croce2020AutoAttack} with attack budget  $\epsilon_\infty=8/255$. Avg. Acc. is the average of clean acc. and robust acc. and used to show overall performance.}
  \label{Tab:transfer_C100_C10}
\end{table}

\begin{table}[t]
  \centering
  \resizebox{1.0\textwidth}{!}{
  \begin{tabular}{@{}lccccccc@{}}
    \toprule
    Model      & \multicolumn{2}{c}{Training Data} & \multicolumn{2}{c}{Test Data}   & Clean Acc. (\%) & Robust Acc. (\%) & Avg. Acc. (\%)\\
               & CIFAR10 & CIFAR100 & CIFAR10 & CIFAR100 & & \\  
    \midrule
    WRN28-10   & & v & & v & 81.66 & 0 & 40.83\\    
    \midrule
    + DiffPure \cite{nie2022DiffPure} & & v & & v & 61.98 & 61.19 & 61.58\\  
    + DISCO \cite{ho2022disco}   & & v & & v & 69.78 & 76.91 & 73.34\\      
    + MAEP     & & v & & v & 73.67 & 76.22 & \textbf{74.95}\\
    \hdashline
    + ScoreOpt-O \cite{zhang2023ScoreOpt} & v & & & v & 57.55 & 42.83 & 50.19\\ 
    + ScoreOpt-N \cite{zhang2023ScoreOpt} & v & & & v & 54.87 & 54.37 & 54.62\\    
    + DiffPure \cite{nie2022DiffPure} & v & & & v & 81.00 & 40.00 & 60.50\\       	
    + DISCO \cite{ho2022disco}   & v & & & v & 72.50 & 69.22 & 70.86\\     
    + MAEP     & v & & & v & 75.37 & 68.75 & \textbf{72.06}\\
    \bottomrule
  \end{tabular}
  }
  \caption{Transferability of adversarial defenses (ScoreOpt, Diffpure, DISCO, and MAEP) from CIFAR10 to CIFAR100 (trained on CIFAR10 but tested on CIFAR100) under WRN28-10 and \AA \ with attack budget  $\epsilon_\infty=8/255$. }
\label{Tab:transfer_C10_C100}
\end{table}

\subsection{Defense Transferability}
\label{sec:Exp:transfer_dataset}

In this section, we examine diffusion-based purifiers \cite{nie2022DiffPure,song2024mimicdiffusion,zhang2023ScoreOpt} under a more challenging setting, where the purifier is applied to a dataset different from the one it was trained. As shown in \Cref{Tab:transfer_C100_C10,Tab:transfer_C10_C100}, these approaches exhibit limited transferability across datasets.

\Cref{Tab:transfer_C100_C10} demonstrates that while DiffPure maintains a small gap between clean and robust accuracy on CIFAR10, its robust accuracy drops significantly from $89.45\%$ to $69.0\%$ when applied to CIFAR100. A similar performance degradation is observed in the reverse transfer setting (\Cref{Tab:transfer_C10_C100}). We highlight DiffPure as a representative method of diffusion-based defenses, as other approaches follow a similar paradigm by modifying the reverse diffusion process.

Crucially, in real-world scenarios, access to a well-trained diffusion model for every potential dataset is often infeasible, and training such models from scratch is computationally expensive—especially for small or diverse datasets. Furthermore, even when the training and testing datasets are closely related, as in the case of CIFAR10 and CIFAR100, diffusion-based defenses \cite{nie2022DiffPure,zhang2023ScoreOpt} suffer from a notable decline in robust accuracy. This highlights a key limitation: a lack of generalization and resilience to slight variations in image distributions.


\subsection{Transferability to High-Resolution Dataset}\label{Sec: More Transfering}

In addition to \Cref{Tab:transfer_C100_C10,Tab:transfer_C10_C100}, we evaluated the transferability of purifiers from a low-resolution dataset to a high-resolution dataset in \Cref{Tab:transfer_imagenet}. 

When transferring from CIFAR-10 to ImageNet, MAEP achieves approximately 75\% clean accuracy, outperforming both DiffPure (68.60\%) and ScoreOpt (68.05\%) at attack budget $\epsilon_\infty=4/255$, even though both baselines are trained directly on ImageNet. 
When attack budget was increased to $\epsilon_\infty=8/255$, MAEP still maintains promising accuracy.
We also evaluated the robust accuracy of DDA, which is specifically designed to maintain classification performance under image corruptions. However, our results show that DDA fails to preserve robustness in this setting.

Additionally, MAEP incurs only a 3\% drop in clean accuracy compared to the original classifier accuracy of 80.85\% without any defense, whereas diffusion-based methods suffer a larger degradation of around 10\%. 
This difference attributes to the fact that diffusion models introduce noises to remove adversarial perturbations, thereby reducing clean accuracy.


\subsection{More Result with Common Color Corruptions}
\label{sec:Exp:Corruption}

\begin{figure}
    \centering
    \includegraphics[width=.9\linewidth]{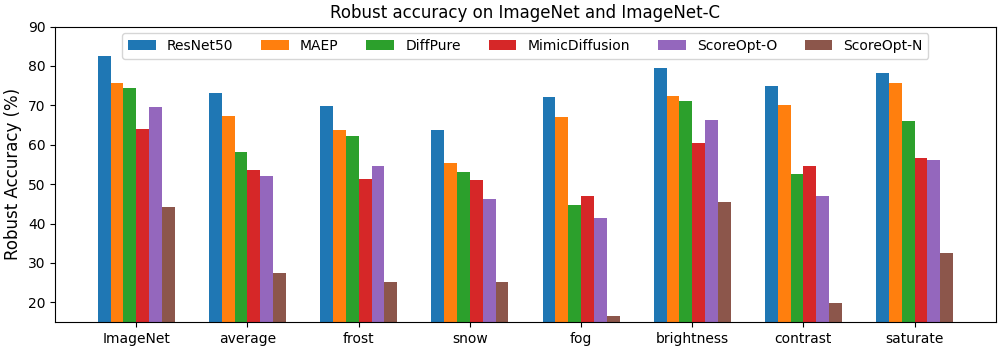}
    \caption{
    Drop in robust accuracy across different purifiers on ImageNet-C. 
    ``ImageNet'' refers to the original, unmodified images, while the other categories represent corrupted versions of ImageNet. 
    The term ``average'' denotes the mean accuracy across various corruptions, excluding the original images. The other terms in the X-asix denote the types of image corruptions.
    }
    \label{fig:corrupt_imagenet_robust_acc}
\end{figure}

We further investigate the impact of additional color-related image corruptions, which are common in real-world settings.
As illustrated in \cref{fig:corrupt_imagenet_robust_acc}, diffusion-based methods exhibit heightened sensitivity to image corruptions. In particular, ScoreOpt demonstrates significant performance degradation. As discussed in Sec. \ref{sec:discrepancy_diffusion}, the stochasticity and approximation involved in ScoreOpt's reverse diffusion process amplify semantic loss, contributing to reduced robustness. Representative examples of purified images are shown in \cref{fig:purified_images}.

While DDA \cite{gao2023dda} is specifically designed to maintain prediction accuracy under corruption, we exclude its results here due to its vulnerability to adversarial perturbations, as evidenced in \cref{Tab:transfer_imagenet}.

\begin{table*}[t]
  \centering   
  \resizebox{1.0\textwidth}{!}{
  \begin{tabular}{@{}l|cc|cc|cccc}  
    \toprule
    Model       & \multicolumn{2}{c}{Train}  & \multicolumn{2}{c}{Test} & Clean Acc. (\%) & Robust Acc. (\%) & Avg. Acc. (\%) & Attacks\\
                & CIFAR10 & ImageNet & CIFAR10 & ImageNet & &  &\\
    
    \midrule
    ResNet50                            & - & v & - & v & 80.85 & 0 & 40.42 & $\epsilon_\infty=4/255$\\
    \hdashline
    + MAEP                              & v & - & - & v & 77.84 & 70.62 & \textbf{74.23} & $\epsilon_\infty=4/255$\\
    + MAEP                              & v & - & - & v & 77.62 & 66.19 & \textbf{71.91} & $\epsilon_\infty=\textbf{8/255}$\\
    + DISCO \cite{ho2022disco}         & v & - & - & v & 76.61 & 69.12 & 72.86 & $\epsilon_\infty=4/255$\\ 
    \cdashline{1-9}
    + MAEP                              & - & v & - & v & 77.97 & 73.94 & \textbf{75.96} & $\epsilon_\infty=4/255$\\   
    + DISCO                             & - & v & - & v    & 77.54 & 70.44 & 73.99 & $\epsilon_\infty=4/255$\\ 
    + DiffPure* \cite{nie2022DiffPure}   & - & v & - & v & 70.01 & 67.11 & 68.60 & $\epsilon_\infty=4/255$\\ 
    + ScoreOpt* \cite{zhang2023ScoreOpt} & - & v & - & v & 70.07 & 66.02 & 68.05 & $\epsilon_\infty=4/255$\\     
    + DDA \cite{gao2023dda} & - & v & - & v & 77.92 & 1.37 & 39.65 & $\epsilon_\infty=4/255$\\ 
    \bottomrule
  \end{tabular}}
  \caption{Transferability of adversarial defenses (DISCO and MAEP) from CIFAR10 to ImageNet (trained on CIFAR10 but tested on ImageNet) under ResNet50 (pre-trained model weight is from official PyTorch) and \AA \cite{croce2020AutoAttack}. sterisk (*) indicates that the results were excerpted from the papers.}
  \label{Tab:transfer_imagenet}
\end{table*}

\section{Conclusion}
\label{sec:Conclusion}

Although diffusion models have demonstrated strong capabilities as adversarial purifiers in prior studies, their limitations remain underexplored. In this paper, we reveal that diffusion-based purification can impair classifier generalization, particularly in scenarios involving color-related variations.
Moreover, we explore the generalization loss in classifiers caused by diffusion models and propose Masked
AutoEncoder Purifier (MAEP), incorporating masked autoencoder (MAE)
and purification loss, as a non-diffusion-based purifier.


\section{Acknowledgement} This work was supported by the National Science and
Technology Council (NSTC) with Grants
NSTC 112-2221-E-001-011-MY2 and 114-2221-E-001 -010 -MY2.

%
%
%

\newpage
\bibliographystyle{splncs04}
\bibliography{reference}

@String(CVPR= {IEEE Conf. Comput. Vis. Pattern Recog.})

@String(ICCV= {Int. Conf. Comput. Vis.})

@String(NIPS= {Adv. Neural Inform. Process. Syst.})

@String(ICLR = {Int. Conf. Learn. Represent.})

@String(AAAI = {AAAI})

@String(CVPR  = {CVPR})

@String(ICCV  = {ICCV})

@String(NIPS  = {NeurIPS})

@String(ICLR  = {ICLR})

@article{WRN,
  title={Wide Residual Networks},
  author={Sergey Zagoruyko and Nikos Komodakis},
  journal={arXiv preprint arXiv:1605.07146},
  year={2017}
}

@article{krizhevsky2009learning,
  title={Learning multiple layers of features from tiny images},
  author={Krizhevsky, Alex and Hinton, Geoffrey and others},
  year={2009},
  publisher={Toronto, ON, Canada}
}

@inproceedings{croce2020AutoAttack,
  title={Reliable evaluation of adversarial robustness with an ensemble of diverse parameter-free attacks},
  author={Croce, Francesco and Hein, Matthias},
  booktitle={ICML},
  pages={2206--2216},
  year={2020},
  organization={PMLR}
}

@article{shafahi2019AdversarialTraining,
  title={Adversarial training for free!},
  author={Shafahi, Ali and Najibi, Mahyar and Ghiasi, Mohammad Amin and Xu, Zheng and Dickerson, John and Studer, Christoph and Davis, Larry S and Taylor, Gavin and Goldstein, Tom},
  journal={NIPS},
  volume={32},
  year={2019}
}

@inproceedings{zhang2019TRADES,
  title={Theoretically principled trade-off between robustness and accuracy},
  author={Zhang, Hongyang and Yu, Yaodong and Jiao, Jiantao and Xing, Eric and El Ghaoui, Laurent and Jordan, Michael},
  booktitle={ICML},
  pages={7472--7482},
  year={2019},
  organization={PMLR}
}

@inproceedings{hsiung2023towards,
  title={Towards compositional adversarial robustness: Generalizing adversarial training to composite semantic perturbations},
  author={Hsiung, Lei and Tsai, Yun-Yun and Chen, Pin-Yu and Ho, Tsung-Yi},
  booktitle={CVPR},
  pages={24658--24667},
  year={2023}
}

@inproceedings{huang2023boosting,
  title={Boosting accuracy and robustness of student models via adaptive adversarial distillation},
  author={Huang, Bo and Chen, Mingyang and Wang, Yi and Lu, Junda and Cheng, Minhao and Wang, Wei},
  booktitle={CVPR},
  pages={24668--24677},
  year={2023}
}

@article{ho2020DDPM,
  title={Denoising diffusion probabilistic models},
  author={Ho, Jonathan and Jain, Ajay and Abbeel, Pieter},
  journal={NIPS},
  volume={33},
  pages={6840--6851},
  year={2020}
}

@inproceedings{kenton2019bert,
  title={BERT: Pre-training of Deep Bidirectional Transformers for Language Understanding},
  author={Kenton, Jacob Devlin Ming-Wei Chang and Toutanova, Lee Kristina},
  booktitle={Proceedings of NAACL-HLT},
  pages={4171--4186},
  year={2019}
}

@inproceedings{he2022MAE,
  title={Masked autoencoders are scalable vision learners},
  author={He, Kaiming and Chen, Xinlei and Xie, Saining and Li, Yanghao and Doll{\'a}r, Piotr and Girshick, Ross},
  booktitle={CVPR},
  pages={16000--16009},
  year={2022}
}

@inproceedings{huang2023FrequencyMAE,
  title={Improving Adversarial Robustness of Masked Autoencoders via Test-time Frequency-domain Prompting},
  author={Huang, Qidong and Dong, Xiaoyi and Chen, Dongdong and Chen, Yinpeng and Yuan, Lu and Hua, Gang and Zhang, Weiming and Yu, Nenghai},
  booktitle={ICCV},
  pages={1600--1610},
  year={2023}
}

@inproceedings{wuDMAE,
  title={Denoising Masked Autoencoders Help Robust Classification},
  author={Wu, QuanLin and Ye, Hang and Gu, Yuntian and Zhang, Huishuai and Wang, Liwei and He, Di},
  booktitle={ICLR},
  year={2023},
}

@article{you2023NoisyImageModeling,
  title={Beyond pretrained features: Noisy image modeling provides adversarial defense},
  author={You, Zunzhi and Liu, Daochang and Han, Bohyung and Xu, Chang},
  journal={NIPS},
  volume={36},
  year={2023}
}

@inproceedings{tsai2023testMAE,
  title={Test-time Detection and Repair of Adversarial Samples via Masked Autoencoder},
  author={Tsai, Yun-Yun and Chao, Ju-Chin and Wen, Albert and Yang, Zhaoyuan and Mao, Chengzhi and Shah, Tapan and Yang, Junfeng},
  booktitle={CVPR Workshops},  
  year={2023}
}

@inproceedings{nie2022DiffPure,
  title={Diffusion Models for Adversarial Purification},
  author={Nie, Weili and Guo, Brandon and Huang, Yujia and Xiao, Chaowei and Vahdat, Arash and Anandkumar, Animashree},
  booktitle={ICML},
  pages={16805--16827},
  year={2022},
  organization={PMLR}
}

@article{zhang2023ScoreOpt,
  title={Enhancing Adversarial Robustness via Score-Based Optimization},
  author={Zhang, Boya and Luo, Weijian and Zhang, Zhihua},
  journal={NIPS},
  volume={36},
  year={2023}
}

@article{ho2022disco,
  title={DISCO: Adversarial defense with local implicit functions},
  author={Ho, Chih-Hui and Vasconcelos, Nuno},
  journal={NIPS},
  volume={35},
  pages={23818--23837},
  year={2022}
}

@inproceedings{chen2019LIIF,
  title={Learning implicit fields for generative shape modeling},
  author={Chen, Zhiqin and Zhang, Hao},
  booktitle={CVPR},
  pages={5939--5948},
  year={2019}
}

@inproceedings{lim2017EDSR,
  title={Enhanced deep residual networks for single image super-resolution},
  author={Lim, Bee and Son, Sanghyun and Kim, Heewon and Nah, Seungjun and Mu Lee, Kyoung},
  booktitle={CVPR workshops},
  pages={136--144},
  year={2017}
}

@inproceedings{song2024mimicdiffusion,
  title={Mimicdiffusion: Purifying adversarial perturbation via mimicking clean diffusion model},
  author={Song, Kaiyu and Lai, Hanjiang and Pan, Yan and Yin, Jian},
  booktitle={CVPR},
  pages={24665--24674},
  year={2024}
}

@inproceedings{gao2023dda,
  title={Back to the source: Diffusion-driven adaptation to test-time corruption},
  author={Gao, Jin and Zhang, Jialing and Liu, Xihui and Darrell, Trevor and Shelhamer, Evan and Wang, Dequan},
  booktitle={CVPR},
  pages={11786--11796},
  year={2023}
}

@inproceedings{somepalli2023digital_forgery,
  title={Diffusion art or digital forgery? investigating data replication in diffusion models},
  author={Somepalli, Gowthami and Singla, Vasu and Goldblum, Micah and Geiping, Jonas and Goldstein, Tom},
  booktitle={CVPR},
  pages={6048--6058},
  year={2023}
}

@inproceedings{madry2018PGD,
  title={Towards Deep Learning Models Resistant to Adversarial Attacks},
  author={Madry, Aleksander and Makelov, Aleksandar and Schmidt, Ludwig and Tsipras, Dimitris and Vladu, Adrian},
  booktitle={ICLR},
  year={2018}
}

@inproceedings{croce2020fab,
  title={Minimally distorted adversarial examples with a fast adaptive boundary attack},
  author={Croce, Francesco and Hein, Matthias},
  booktitle={ICML},
  pages={2196--2205},
  year={2020},
  organization={PMLR}
}

@article{goodfellow2015FGSM,
  title={EXPLAINING AND HARNESSING ADVERSARIAL EXAMPLES},
  author={Goodfellow, Ian J and Shlens, Jonathon and Szegedy, Christian},
  journal={stat},
  volume={1050},
  pages={20},
  year={2015}
}

@incollection{kurakin2018BIM,
  title={Adversarial examples in the physical world},
  author={Kurakin, Alexey and Goodfellow, Ian J and Bengio, Samy},
  booktitle={Artificial intelligence safety and security},
  pages={99--112},
  year={2018},
  publisher={Chapman and Hall/CRC}
}

@inproceedings{athalye2018BPDA,
  title={Obfuscated gradients give a false sense of security: Circumventing defenses to adversarial examples},
  author={Athalye, Anish and Carlini, Nicholas and Wagner, David},
  booktitle={ICML},
  pages={274--283},
  year={2018},
  organization={PMLR}
}

@inproceedings{rombach2022LatentDiffusion,
  title={High-resolution image synthesis with latent diffusion models},
  author={Rombach, Robin and Blattmann, Andreas and Lorenz, Dominik and Esser, Patrick and Ommer, Bj{\"o}rn},
  booktitle={CVPR},
  pages={10684--10695},
  year={2022}
}

@inproceedings{radford2021CLIP,
  title={Learning transferable visual models from natural language supervision},
  author={Radford, Alec and Kim, Jong Wook and Hallacy, Chris and Ramesh, Aditya and Goh, Gabriel and Agarwal, Sandhini and Sastry, Girish and Askell, Amanda and Mishkin, Pamela and Clark, Jack and others},
  booktitle={ICML},
  pages={8748--8763},
  year={2021},
  organization={PMLR}
}

@inproceedings{dosovitskiy2020ViT,
  title={An Image is Worth 16x16 Words: Transformers for Image Recognition at Scale},
  author={Dosovitskiy, Alexey and Beyer, Lucas and Kolesnikov, Alexander and Weissenborn, Dirk and Zhai, Xiaohua and Unterthiner, Thomas and Dehghani, Mostafa and Minderer, Matthias and Heigold, Georg and Gelly, Sylvain and others},
  booktitle={ICLR},
  year={2020}
}

@inproceedings{alfarra2022AntiAdversaries,
  title={Combating adversaries with anti-adversaries},
  author={Alfarra, Motasem and P{\'e}rez, Juan C and Thabet, Ali and Bibi, Adel and Torr, Philip HS and Ghanem, Bernard},
  booktitle={AAAI},
  volume={36},
  number={6},
  pages={5992--6000},
  year={2022}
}

@article{krizhevsky2009CIFAR10,
  title={Learning multiple layers of features from tiny images},
  author={Krizhevsky, Alex and others},
  year={2009}
}

@inproceedings{deng2009imagenet,
  title={Imagenet: A large-scale hierarchical image database},
  author={Deng, Jia and Dong, Wei and Socher, Richard and Li, Li-Jia and Li, Kai and Fei-Fei, Li},
  booktitle={CVPR},
  pages={248--255},
  year={2009},
  organization={Ieee}
}

@article{hendrycks2019imagenetC,
  title={Benchmarking neural network robustness to common corruptions and perturbations},
  author={Hendrycks, Dan and Dietterich, Thomas},
  journal={arXiv preprint arXiv:1903.12261},
  year={2019}
}

@article{reinhard2001ColorTransfer,
  title={Color transfer between images},
  author={Reinhard, Erik and Adhikhmin, Michael and Gooch, Bruce and Shirley, Peter},
  journal={IEEE Computer graphics and applications},
  volume={21},
  number={5},
  pages={34--41},
  year={2001},
  publisher={IEEE}
}

@inproceedings{zagoruyko2016WRN,
  title={Wide Residual Networks},
  author={Zagoruyko, Sergey and Komodakis, Nikos},
  booktitle={British Machine Vision Conference 2016},
  year={2016},
  organization={British Machine Vision Association}
}

@article{wu2020adversarial,
  title={Adversarial weight perturbation helps robust generalization},
  author={Wu, Dongxian and Xia, Shu-Tao and Wang, Yisen},
  journal={NIPS},
  volume={33},
  pages={2958--2969},
  year={2020}
}

@article{shi2021online,
  title={Online adversarial purification based on self-supervision},
  author={Shi, Changhao and Holtz, Chester and Mishne, Gal},
  journal={arXiv preprint arXiv:2101.09387},
  year={2021}
}

@inproceedings{hill2020stochastic,
  title={Stochastic Security: Adversarial Defense Using Long-Run Dynamics of Energy-Based Models},
  author={Hill, Mitch and Mitchell, Jonathan Craig and Zhu, Song-Chun},
  booktitle={ICLR},
  year={2020}
}

@inproceedings{yoon2021adversarial,
  title={Adversarial purification with score-based generative models},
  author={Yoon, Jongmin and Hwang, Sung Ju and Lee, Juho},
  booktitle={ICML},
  pages={12062--12072},
  year={2021},
  organization={PMLR}
}

@article{rebuffiNeurIPS2021,
  title={Data augmentation can improve robustness},
  author={Rebuffi, Sylvestre-Alvise and Gowal, Sven and Calian, Dan Andrei and Stimberg, Florian and Wiles, Olivia and Mann, Timothy A},
  journal={NIPS},
  volume={34},
  pages={29935--29948},
  year={2021}
}

@inproceedings{wangICML2023,
  title={Better diffusion models further improve adversarial training},
  author={Wang, Zekai and Pang, Tianyu and Du, Chao and Lin, Min and Liu, Weiwei and Yan, Shuicheng},
  booktitle={ICML},
  pages={36246--36263},
  year={2023},
  organization={PMLR}
}

@article{cuiarXiv2023,
  title={Decoupled kullback-leibler divergence loss},
  author={Cui, Jiequan and Tian, Zhuotao and Zhong, Zhisheng and Qi, Xiaojuan and Yu, Bei and Zhang, Hanwang},
  journal={arXiv preprint arXiv:2305.13948},
  year={2023}
}

\section*{Appendix}



\section{More Objective Function Designs}\label{Sec: Objective Functions}
\textbf{1) Reconstruction Loss.}
Intuitively, to enhance both the clean accuracy and robust accuracy of an NN model, the purified/reconstructed image should closely resemble its clean version by minimizing the loss as:
\begin{align}  
L = D(\mathcal{P}(x_a), x) + D(\mathcal{P}(x), x),
\label{Eq:Loss_adv_clean}
\end{align}
where the first term denotes the distance between the purified adversarial image and its corresponding clean image, and the second term measures the distance between the purified clean image and true clean one.

\textbf{2)Masked Language Modeling (MLM).}
We leverage MLM directly to train the purifier through a two-step process. Initially, it aims to learn adversarial embeddings during the pre-training stage, as indicated in \cref{Eq:MAE}, and subsequently finetunes to purify an adversarial image. The loss function is described as follows with respect to the two-step process:

\begin{align}  
L_{pre-train}=L_{MAE}(x_a,x)=\|({\bf 1}-M) \odot x - ({\bf 1}-M) \odot (g \circ f (M\odot\ x_a))\|.
\label{Eq:Loss_MLM_pre-trin}
\end{align}

\begin{align}  
L_{finetune}=D(\mathcal{P}_{MAE}(x_a), x).
\label{Eq:Loss_MLM_finetune}
\end{align}

\textbf{3) TRADES \cite{zhang2019TRADES}.}
TRADES proposed to train a robust classifier with the loss function defined as:
\begin{align}  
L = CrossEntropy(c(x), y) + KL(c(x), c(x_a))/ \lambda,
\label{Eq:Loss_TRADES}
\end{align}
where the first term maintains the clean accuracy while the second term focuses on improving the robust accuracy by making logits of adversarial sample similar to those of clean acccuracy, and $c$ is a classifier.

\textbf{3.1) TRADES in pixel domain.}
To replicate the success of TRADES in adversarial training, we adapt its concept from training a robust classifier to training an adversarial purifier. The main difference is that the purifier needs to process the image instead of class prediction. Therefore, we replace the cross-entropy loss and KL divergence loss in \cref{Eq:Loss_TRADES} with a reconstruction loss in the image pixel domain to meet the purifier's requirement as:
\begin{align}  
L = D(\mathcal{P}(x), x) + D(\mathcal{P}(x), \mathcal{P}(x_a))/ \lambda,
\label{Eq:Loss_TRADES_image}
\end{align}
where the first term maintains the purified clean image quality and the second term tries to purify the adversarial image by mimicking the clean image in a sense similar to KL divergence loss in \cref{Eq:Loss_TRADES}.

\textbf{3.2) TRADES in latent domain.}
Unlike the methods proposed to concentrate on the image pixel domain, several works, such as Latent Diffusion \cite{rombach2022LatentDiffusion} and CLIP \cite{radford2021CLIP}, have achieved notable success by processing image latent representations. In our approach, as indicated in Eq. (\ref{Eq:Loss_TRADES_latent}) below, we maintain  clean image purification (1st term) while enforcing constraints on adversarial perturbations within the latent space (2nd term) as:
\begin{align}  
L = D(\mathcal{P}(x), x) + D(f(x), f(x_a))/ \lambda.
\label{Eq:Loss_TRADES_latent}
\end{align}

In \Cref{Tab:objective_function}, we provide a comparison of the performance of all loss designs discussed here to verify the design of MAEP. We have the following observations: (1) Although reconstruction loss concurrently learns the reconstruction of both clean and adversarial images, its performance falls short of DISCO, which concentrates solely on reconstructing adversarial images. (2) MLM and DISCO are closely associated with MAEP. Directly applying MLM appears ineffective, while MAEP demonstrates performance enhancement over DISCO. (3) Exploiting the concept of TRADES does not aid in learning an adversarial purifier. Thus, our validation shows that MAEP significantly outperforms other approaches.


\begin{table}[h]
  \centering    
  \begin{tabular}{@{}l|ccc}  
    \toprule
    Defenses & Clean Acc. (\%) & Robust Acc. (\%) & Avg. Acc.\\
    \hline
    \midrule
    WRN28-10       & 94.78 & 0 & 47.39\\
    \hdashline
    \ + DISCO \cite{ho2022disco} & 89.26  & 85.33 & 87.29\\
    \hdashline
    \ + Reconstruction (Eq. (\ref{Eq:Loss_adv_clean}))     & 94.74 & 82.73 & 88.73\\
    \ + TRADES (pixel, Eq. (\ref{Eq:Loss_TRADES_image}))   & 94.75 & 0.85  & 47.81\\
    \ + TRADES (latent, Eq. (\ref{Eq:Loss_TRADES_latent})) & 94.64 & 38.16 & 66.40\\
    \ + MLM \cite{he2022MAE}                               & 92.85 & 61.46 & 77.15\\
    \hdashline
    \ + MAEP & 92.30 & 88.73 & \textbf{90.52}\\
    \bottomrule
  \end{tabular}
  \caption{Performance comparison of different objective functions on CIFAR10 dataset under \AA \ with attack budget  $\epsilon_\infty=8/255$.}
  \label{Tab:objective_function}
\end{table}

\end{document}